\documentclass[letterpaper, 10 pt, conference]{ieeeconf}
\IEEEoverridecommandlockouts   
\usepackage{cite}                       
\usepackage{amsmath,amssymb,amsfonts}
\usepackage{graphicx}
\usepackage{textcomp}
\usepackage{xcolor}
\usepackage{booktabs}
\usepackage{array}
\usepackage{paralist}                   
\usepackage{balance}                    
\usepackage[hidelinks,breaklinks=true]{hyperref}
\hypersetup{pdfauthor={},pdftitle={},pdfsubject={},pdfkeywords={}}
\usepackage[capitalize]{cleveref}       
\usepackage{microtype}                  

\graphicspath{{figures/}}

\crefname{figure}{Fig.}{Figs.}
\Crefname{figure}{Fig.}{Figs.}
\crefname{table}{Table}{Tables}
\Crefname{table}{Table}{Tables}
\crefname{section}{Section}{Sections}
\Crefname{section}{Section}{Sections}
\crefname{equation}{}{}
\Crefname{equation}{}{}

\makeatletter
\renewcommand\paragraph{\@startsection{paragraph}{4}{\z@}%
  {1.4ex \@plus .4ex \@minus .2ex}%
  {-0.55em}%
  {\normalfont\normalsize\bfseries}}
\def\@IEEEsectpunct{\ \,}
\makeatother

\newif\ifanonymous

\title{\LARGE \bf
Agentic Real2Sim: Physics-based World Modeling\\ with Vision-Language Agents
}

\ifanonymous
\author{Anonymous Author(s)}
\else
\author{%
Guanxiong Chen$^{1, 6,*,\#}$, Qianjun Xia$^{1,*}$, Jiawei Peng$^{2,*}$, Heng Zhang$^{1,*}$, Pengyu Jing$^{6, *}$, Bole Ma$^{3,*}$, \\
Justin Qian$^{1}$, Yixian Cheng$^{1}$, Ziyi Jiao$^{1}$, Bingyang Zhou$^{6}$, Yiduo Qu$^{7}$, Luoxin Ye$^{2}$, Kaifeng Zhang$^{4}$, \\ Kunyi Wang$^{1}$, Weijia Zeng$^{1}$, Yunuo Chen$^{5}$, Pengzhi Yang$^{6}$, Ziqiu Zeng$^{6}$, Siyuan Luo$^{6}$, \\
Huamin Wang$^{8}$, Chao Liu$^{1}$, Alan Yuille$^{2}$, Fan Shi$^{6}$, \\
Changxi Zheng$^{4}$, Yunzhu Li$^{4}$, Chenfanfu Jiang$^{5}$, Peter Yichen Chen$^{1}$%
\thanks{$^{*}$Equal technical contribution.\quad $^{\#}$Project lead.}%
\thanks{$^{1}$University of British Columbia.
        $^{2}$Johns Hopkins University.
        $^{3}$Erlangen National High Performance Computing Center (NHR@FAU),
        Friedrich-Alexander-Universit\"at Erlangen-N\"urnberg.
        $^{4}$Columbia University.
        $^{5}$University of California, Los Angeles.
        $^{6}$National University of Singapore.
        $^{7}$University of Cambridge.
        $^{8}$Style3D.}%
}
\fi

\makeatletter
\newcommand{\titleteaser}{%
  \begin{minipage}{\textwidth}
    \centering
    \includegraphics[width=0.95\linewidth]{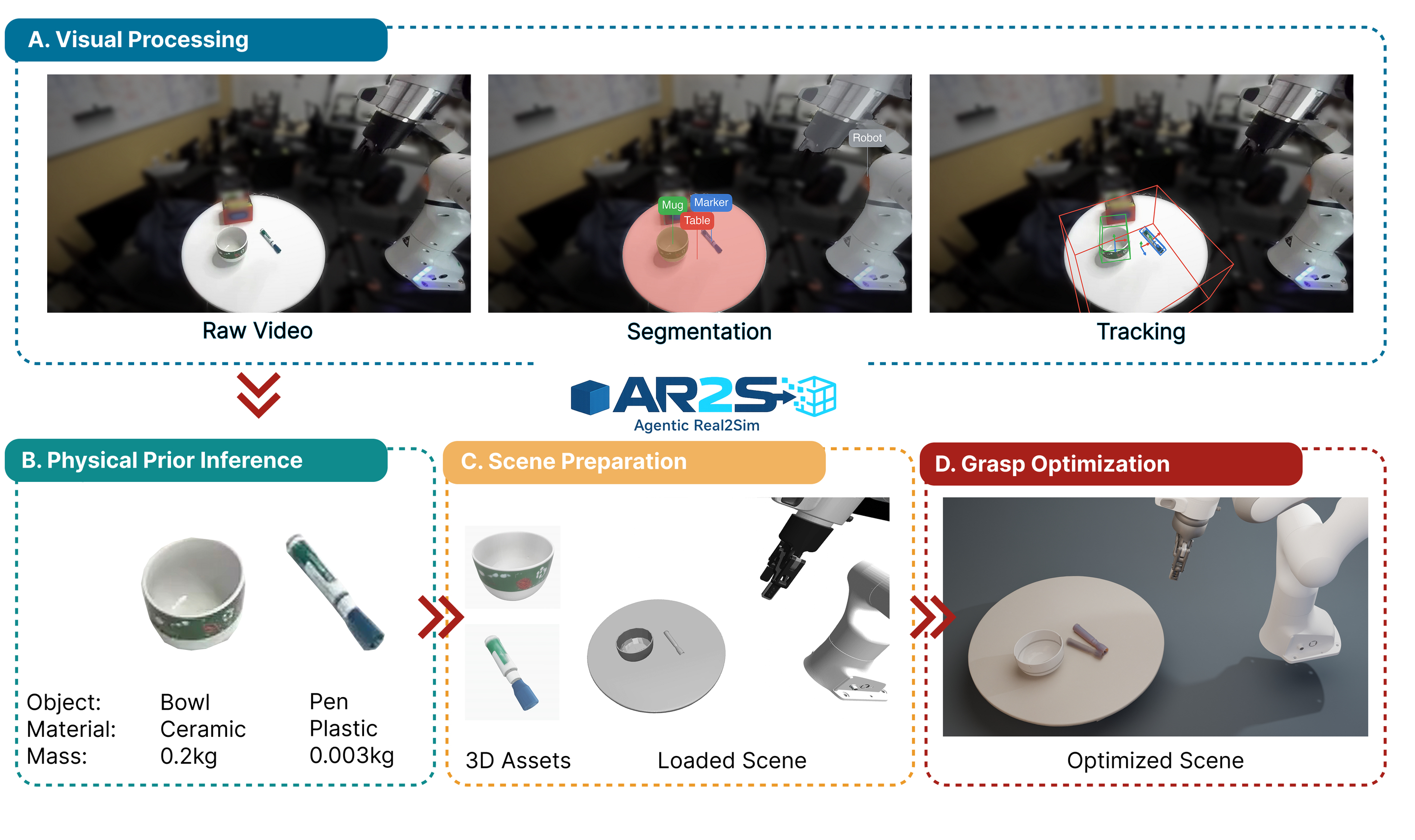}
    \par
    \def\@captype{figure}%
    \refstepcounter{figure}%
    \@makecaption{\fnum@figure}{\textbf{Agentic Real2Sim architecture.} A recorded DROID episode is converted into a simulatable digital twin through four linked agents: The visual processing agent extracts segmentation and depth masks, geometries and object poses from raw video. The physical-prior inference agent attaches object identity, material class, mass hints, and other constitutive properties needed by the simulator. The scene preparation agent initializes a scene with the robot's initial state, cameras, and reconstructed geometry, and places objects of interest in the correct locations. Simulator-in-the-loop grasp optimization further optimizes object placements to enable successful grasping. The same episode contract supplies the inputs and feedback channels used by the rigid, deformable, and humanoid adapters.}%
    \label{fig:teaser}%
  \end{minipage}%
}
\makeatother
\IEEEaftertitletext{\titleteaser}

\begin{document}
\bstctlcite{IEEEexample:BSTcontrol}
\maketitle

\begin{figure*}[t]
  \centering
  \includegraphics[width=.95\linewidth]{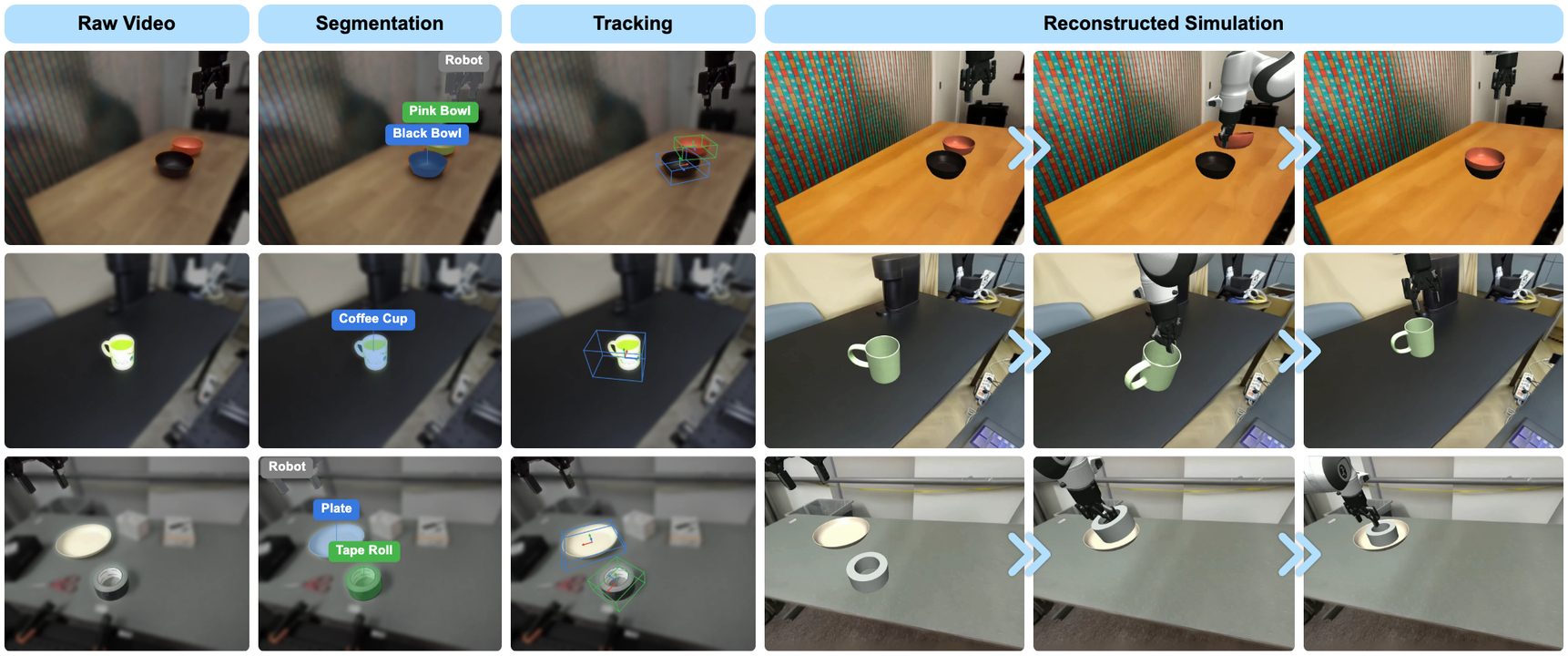}
  \caption{\textbf{DROID episode conversion at scale.} Three representative episodes from \textbf{DROID-500}, each pairing the episode's real-world observation with its synthetic twin. Intermediate pipeline artifacts are exposed for selected scenes: object segmentation and FoundationPose pose-tracking overlays.}
  \label{fig:droid-qual}
\end{figure*}

\begin{abstract}
Real-to-sim conversion for robotic interaction with objects remains labor-intensive because it requires more than visual reconstruction: a streamlined real2sim process must recover scene geometries and object states, infer physical parameters, and assemble actors, objects, cameras, poses, and trajectories into a runnable physical simulation. Today this process still depends on brittle workflow glue across visual perception tools and simulators: manual tuning of visual foundation models, mesh cleanup, coordinate frame alignments, etc. We introduce \textit{Agentic Real2Sim}, a framework for generalized physical world modeling with vision-language agents that converts a real-world recording of object-robot interaction into a simulatable episodic twin, and connects the resulting twin to downstream policy fine-tuning and evaluation. We evaluate Agentic Real2Sim on rigid-object manipulation, deformable-object interaction, and humanoid motion scenes, spanning domains that are usually handled by separate Real2Sim pipelines. The framework's agentic decisions can be driven by an open-weight VLM backend at a small fraction of the cost of frontier models, while attaining a comparable conversion success rate. The framework further supports custom scene conversion, fine-tuning of a pretrained policy with data generated from converted episodes, and works effectively as a surrogate for real-world policy evaluation. The project site, including code is available at \url{https://agentic-real2sim.github.io/}.
\end{abstract}


\section{Introduction}
\label{sec:intro}

Large real-world robot datasets and increasingly capable simulators have made it tempting to treat real-to-sim conversion as a solved plumbing problem: collect observations, reconstruct assets, drop them into a physics engine, and train or evaluate policies. Yet in practice, physical interaction episodes resist this simplification: a manipulation recording contains a rich set of entities--- cameras, manipulated objects with their physical parameters, kinematic and dynamic states, and trajectories that dictate how a robot behaves across space and time. Accumulating these elements into a \textit{simulatable} twin episode still requires a significant amount of manual labor: configuring visual foundation models, cleaning up meshes, aligning coordinate frames, and tuning physical simulators. Further, automating these steps still requires human-made decisions, such as identifying the names of objects of interest and selecting frames for object identification.

Recent works have made strong progress on parts of this problem. Automated Real2Sim frameworks recover simulation-ready assets and physical parameters from robotic interaction~\cite{Pfaff_2025}, while Real2Sim2Real frameworks align visual and dynamic properties for manipulation policy development~\cite{fan2025twinaligner, Jiang_2025, Chen_2026}. Meanwhile, state-of-the-art VLM-based agentic data-generation frameworks use the generate-critic-improve paradigm to build simulation-ready scenes~\cite{ma2026lychsim, wang2026physcensis, Zook_2025, pfaff2026scenesmith} or object-level assets~\cite{zhou2026articraft}. These lines of work point to a missing capability: scalable conversion of recorded physical interaction episodes into physically simulatable digital twins. Given a real-world recording of an interaction, can a framework produce a simulatable artifact that replays the episode, supports closed-loop policy evaluation, and generates more data for policy fine-tuning and evaluation?

We frame this problem as \emph{Agentic Real2Sim}: Given a recorded physical interaction episode, automatically produces a physically realistic digital twin that preserves the actors, objects, trajectories, and physical parameters, then uses the resulting twin for downstream policy evaluation and fine-tuning---see Fig.~\ref{fig:teaser}. Our contributions are threefold:
\begin{inparaenum}[(1)]
\item We introduce a reproducible agentic real2sim framework to convert robot manipulation episodes~\cite{Khazatsky_2024} into MuJoCo-simulated episodic twins~\cite{Todorov_2012}. The system transforms recorded robot-object interactions into simulatable artifacts through visual processing, physical-prior inference, scene preparation, and simulator-in-the-loop refinement. Moreover, we generalize the framework beyond the DROID manipulation dataset, by enabling the same conversion process for deformable manipulation in a PhysTwin-style setting~\cite{Jiang_2025}, and for humanoid motion with BFM-Zero-style motion context~\cite{li2025bfmzeropromptablebehavioralfoundation}.
\item We demonstrate that the framework supports interchangeable VLM backends, without relying on a single frontier VLM. We demonstrate that with our framework, an open 31B VLM achieves comparable replay-success rate to frontier VLMs on a dataset of 100 episodes, while reducing the cost by up to $31.4\times$.
\item We demonstrate downstream robotic applications with custom scene conversion, data generation through augmentation, and use of generated episodes as a substitute for real-world evaluation.
\end{inparaenum}

\section{Related Work}
\label{sec:related}

\paragraph{Automated real-to-sim alignment.}
Real2Sim framework such as Scalable Real2Sim and TwinAligner~\cite{Pfaff_2025, fan2025twinaligner} have shifted from manually authored simulator scenes toward automated geometry and parameter recovery. These works establish that reconstructing photorealistic scenes alone is insufficient for robotic learning: simulators used for manipulation must also recover dynamics, contacts, and other task-relevant context for generalizable action planning~\cite{liu2026simpact}. Agentic Real2Sim shares this emphasis on physical alignment, but differs in its unit of conversion. Rather than scanning isolated objects or constructing a scene for a new policy-training setup, it converts recorded interaction episodes into simulatable twins that preserve the observed actors, manipulated objects, trajectories, contacts, and ensure visual alignment.

\paragraph{VLM-driven asset generation.} Recent works leverage state-of-the-art visual-language models for generating visually realistic and physically plausible assets for downstream applications. While earlier frameworks primarily rely on one-off VLM API calls embedded in larger scene-reconstruction and task-generation pipelines~\cite{Yang_2024, Katara_2024, Zook_2025, ranawaka2026simfoundry}, more recent works such as SceneSmith~\cite{pfaff2026scenesmith}, SceneWeaver~\cite{yang2026sceneweaver}, PhyScensis~\cite{wang2026physcensis}, SimWorld Studio~\cite{kang2026simworld}, and LychSim~\cite{ma2026lychsim} have shown the strong potential of agentic VLM pipelines in reconstructing realistic scenes from real-world data, or building articulated assets through agentic code and testing loops~\cite{zhou2026articraft}. These agentic frameworks enable VLMs to operate directly over intermediate generation artifacts, using structured, artifact-level feedback to identify omissions or implausibilities, and to iteratively refine subsequent generation decisions. Our work follows this paradigm of using VLM-driven agents for synthetic scene generation and real-to-sim construction, yet places greater emphasis on reconstructing full manipulation behaviors from real-world recordings, as well as generating physical parameters, in addition to placing assets appropriately to recreate visually realistic scenes.


\paragraph{Visual processing and simulation tools.}
Our pipeline also relies on replaceable visual and simulation components. SAM 3 supplies open-vocabulary segmentation for selecting and tracking relevant objects or regions~\cite{carion2025sam}. SAM 3D is used as a geometry-extraction component, not as the object-discovery mechanism, by recovering 3D object assets from image evidence~\cite{chen2025sam}. FoundationStereo provides stereo depth estimation from rectified camera pairs~\cite{wen2025stereo}, and FoundationPose provides 6D object pose estimation and tracking for novel objects~\cite{Wen_2024}. Deformable simulation tools such as PhysTwin and EMPM further show how visual evidence can be coupled with simulator rollouts and parameter optimization to recover material or connectivity parameters~\cite{Jiang_2025,Chen_2026}. These components are important building blocks, but they are not the paper's main contribution. Agentic Real2Sim contributes the episode contract and alignment loop that compose visual extraction, simulator execution, critic feedback, and domain-specific repair into executable episode twins.

\paragraph{Downstream use of synthetic episodes.}
Simulation-ready episodes enable physical queries beyond visual reconstruction. Prior works on demonstration amplification and simulation data generation study how converted or synthesized demonstrations can support skill learning and deployment~\cite{garrett2024skillmimicgen, wan2025lodestar}. Real2Sim2Real platforms such as~\cite{zhang2025real, zhang2026robosnap} use reconstructed objects and scenes as a test bed for policies: policies trained can be first evaluated in the synthetic environment before being evaluated in the real world, which can be time-consuming and expensive. Our framework can be useful for a multitude of robotics downstream tasks, e.g., transforming real-world teleoperation episodes into digital twins on a custom data collection platform, generating more synthetic data to fine-tune a DROID-pretrained $\pi_{0.5}$ policy~\cite{physicalintelligence2025pi05}, or being used as a test bed to evaluate policies fine-tuned solely on real-world data.

\section{Method}
\label{sec:method}


\subsection{System Overview}
\label{sec:method-overview}

\Cref{fig:teaser} summarizes the architecture through a rigid-object manipulation example. The \emph{visual processing agent} identifies task-relevant objects, the manipulated object, and their support relations, then selects clean keyframes and segments the actors and objects. It rejects poor masks, reconstructs and scales meshes, tracks 6-DoF object poses, checks scale and tracking quality, and emits a canonical episode folder. The \emph{physical-prior inference agent} estimates each object's identity, material, mass, and contact attributes for simulation, sharing the spirit of NeRF2Physics~\cite{zhai2024physical}. The \emph{scene preparation agent} selects the primary camera during preprocessing, calibrates the cameras, aligns the robot base, objects, trajectories, and ground reference, and loads the assembled scene into MuJoCo. Finally, the \emph{simulator-in-the-loop grasp optimization agent} tests small shifts of the manipulated object and retains the placement that produces a successful grasp. The tools and skills exposed to these agents are listed in \Cref{tab:tools}. The system separates agentic decisions from low-level tools: agents resolve ambiguous evidence, choose episode adapters, and decide how to refine a result, while specialized tools handle extraction, rendering, pose optimization, and grasp optimization. This design is consistent with prior agentic generation systems such as SimWorld Studio and ArtiCraft~\cite{kang2026simworld,zhou2026articraft}. We use LangChain as our agentic harness, also used by prior agentic research frameworks~\cite{bran2023chemcrow}, for model interfaces, tool calls, and agent orchestration. This infrastructural setup allows us to clearly separate different layers--- prompts and skills vs. tools, and easily switch between VLM backends, thereby focusing on episode reconstruction and real-sim alignment.

\subsection{Converting DROID Episodes}
\label{sec:method-rigid}

\begin{table}[t]
  \centering
  \caption{Tools and skills exposed to the DROID conversion agents.}
  \label{tab:tools}
  \footnotesize
  \setlength{\tabcolsep}{3pt}
  \renewcommand{\arraystretch}{1.12}
  \begin{tabular}{@{}
      >{\raggedright\arraybackslash}p{0.27\linewidth}
      >{\raggedright\arraybackslash}p{0.69\linewidth}@{}}
    \toprule
    \textbf{Agent} & \textbf{Tools and skills} \\
    \midrule
    Visual processing agent
      & \texttt{object\_discovery}, \texttt{object\_relevance}, \texttt{pickup\_object}, and \texttt{support\_tree}; \texttt{segment} \textit{(SAM~3)} with keyframe selection and a mask critic; \texttt{mesh\_recover} \textit{(SAM~3D)} and \texttt{mesh\_scale} with \texttt{scale\_critic}; \texttt{pose\_tracking} \textit{(FoundationPose)} with \texttt{tracking\_critic}. \\
    \addlinespace[3pt]
    Physical-prior inference agent
      & \texttt{material\_classify}. \\
    \addlinespace[3pt]
    Scene preparation agent
      & \texttt{camera\_select}, \texttt{ground\_ref}, \texttt{calibration}, and base-pose optimization. \\
    \addlinespace[3pt]
    Simulator-in-the-loop grasp optimization agent
      & \texttt{grasp\_sweep}. \\
    \bottomrule
  \end{tabular}
\end{table}

Our framework primarily focuses on DROID-style robot-object interactions~\cite{Khazatsky_2024}. The input contains synchronized RGBD streams, calibration data, a robot trajectory, and optionally a task description. The visual pipeline first selects the primary external camera from the synchronized streams, then extracts frames from its recording and builds a rectified video. A VLM-driven object discovery node proposes scene entities, after which a keyframe selector chooses a frame for segmentation. A mask critic can reject poor segmentation and route the graph back to keyframe selection with a bounded retry budget. Accepted masks feed mesh recovery, while full-image depth reconstruction~\cite{wen2025stereo} further enables mesh scaling and FoundationPose-style object tracking~\cite{Wen_2024}. A tracking critic records pose-quality judgments and can request a new initialization frame. Finally, two VLM queries identify the object the robot manipulates and the scene entity that serves as the ground reference.

The resolved visual state is written into an episode folder containing object meshes, scales, pose tracks, robot trajectory, camera metadata, first-frame observations, masks, and task semantics. Next, the scene preparation agent proceeds: it invokes a deterministic calibration substage, optimizes the robot base pose against the robot mask, and estimates the ground plane's orientation and offset beneath the ground reference chosen upstream. The agent then loads the calibrated scene into MuJoCo~\cite{Todorov_2012}, and makes a decision: either it performs a deterministic sweep over object shift parameters, or it calls a VLM-assisted loop that proposes refinements from replay evidence. The sweep path evaluates a batch of candidate shifts against contact and grasp outcomes; the loop path exposes rendered keyframes and structured replay summaries to the agent before each refinement.

\paragraph{Photorealistic rendering.} Agentic Real2Sim combines an interactive foreground with a re-renderable visual context, following the layered rendering strategy of RoboSnap~\cite{zhang2026robosnap}: manipulated objects in the foreground reuse the appearance textures generated by SAM~3D, while the background rendering routine completes regions occluded by the interaction area, reconstructs the static scene as a Gaussian-splatting representation, and registers it to the episode camera and world frames. For a queried camera, the simulator renders foreground color, depth, and alpha channels, and the Gaussian-splatting renderer produces the background view from the same camera pose. Depth-aware compositing combines these layers into photorealistic frames used as policy observations and qualitative real-to-sim comparisons.

\subsection{Converting PhysTwin and BFM-Zero Episodes}
\label{sec:method-deformable-humanoid}

We extend the same episode contract beyond rigid DROID manipulation. For deformables, following the PhysTwin/EMPM setting~\cite{Jiang_2025,Chen_2026}, the framework replaces rigid object pose tracking with tracked geometry, point, particle, spring, or material state and uses simulator rollouts to check whether recovered parameters reproduce observed deformation. For converting humanoid motion from \cite{li2025bfmzeropromptablebehavioralfoundation}, the framework replaces object-centric scene preparation with motion-context retrieval, embodiment-specific initialization, and closed-loop replay against retargeted reference motion, using motion priors from the humanoid/video domain~\cite{harvey2020robust}. For both deformable and humanoid motion conversion tasks, the framework uses the same agentic workflow, while only exposing domain-specific skills and tools, without forcing all episodes into rigid-body representation.

\section{Experiments}
\label{sec:experiments}

Prior works have studied critic-guided refinement through critic removal in SceneSmith~\cite{pfaff2026scenesmith} and reflection ablations in SceneWeaver~\cite{yang2026sceneweaver}. We adopt this existing design and focus our evaluation on episodic conversion capabilities, VLM cost, and downstream uses. We therefore do not include a separate actor--critic ablation.

\subsection{Evaluation Metrics}
\paragraph{VLM as judge.} We evaluate the rigid DROID conversions with replay success, an episode-level metric that measures consistency between the simulated episode and the real episode. Specifically, following the paradigm adopted in~\cite{cao2026physx}, we compute the metric following a structured, VLM-based process: an evaluator first prepares a fixed set of real and simulated keyframes with the labels \texttt{start}, \texttt{middle\_1}, \texttt{middle\_2}, and \texttt{end}. Next, the evaluator selects reconstruction candidates deterministically from the latest grasp sweep: a sample is eligible only if the probe marks it as grasped, its replay video exists, and its lift and displacement statistics are finite and within broad sanity bounds. Among the eligible samples, the evaluator sends at most five candidates to the judges, ranked by peak object displacement with sample id as the deterministic tie-breaker. We use three frontier models from different labs as judges: GPT-5.5 (OpenAI), Claude Opus 4.7 (Anthropic), and Gemini 3.5 Flash (Google); judges are deliberately chosen to avoid overlap with the pipeline backend. Each judge independently scores each candidate out of $10$, subtracting up to $4$ points for identifying the wrong target object, $3$ for a wrong final object location, $2$ for a wrong action performed, and $1$ for a wrong final gripper location. Starting-pose drift is recorded as context rather than directly subtracted; scores of $8$ or higher count as pass, $7$ as partial, and $6$ or lower as fail. Each judge acts as a success finder by nominating its own highest-scored candidate. We define $r_e\in\{0,1\}$ as the episode replay-success indicator, with $r_e=1$ if at least one of the three judges assigns a best-candidate score of at least $8$ out of $10$; otherwise $r_e=0$. Episodes without a valid judge record are also assigned $r_e=0$. The reported replay score for the episode is the maximum of the three judge-best scores.

\paragraph{Visual and Pose Metrics.} We supplement the VLM-based evaluation with deterministic visual- and pose-based metrics: full-frame PSNR and SSIM for visual fidelity, and ADD-S in mm for pose trajectory accuracy. These metrics are applied to episodes that produce evaluable artifacts.

\subsection{DROID Episode Conversion at Scale}
\label{sec:exp-droid-scale}

We evaluate the rigid object manipulation setting on a randomly selected set of $500$ manipulation episodes sampled to span objects, camera viewpoints, occlusion patterns, and manipulation verbs (pick / place / take / remove). All sampled episodes remain in the replay-success denominator, including runs that stop before producing a valid replay record.

\paragraph{Quantitative.} Using Gemma 4 31B as the VLM backend, Agentic Real2Sim produces 242 ($48.4\%$) successful, 28 ($5.6\%$) partial, and 230 ($46.0\%$) failed replay outcomes. Among 349 evaluable episodes, we find a mean PSNR of 14.069, SSIM of 0.650, and ADD-S of 107 mm.

\begin{figure}[t]
  \centering
  \includegraphics[width=\linewidth]{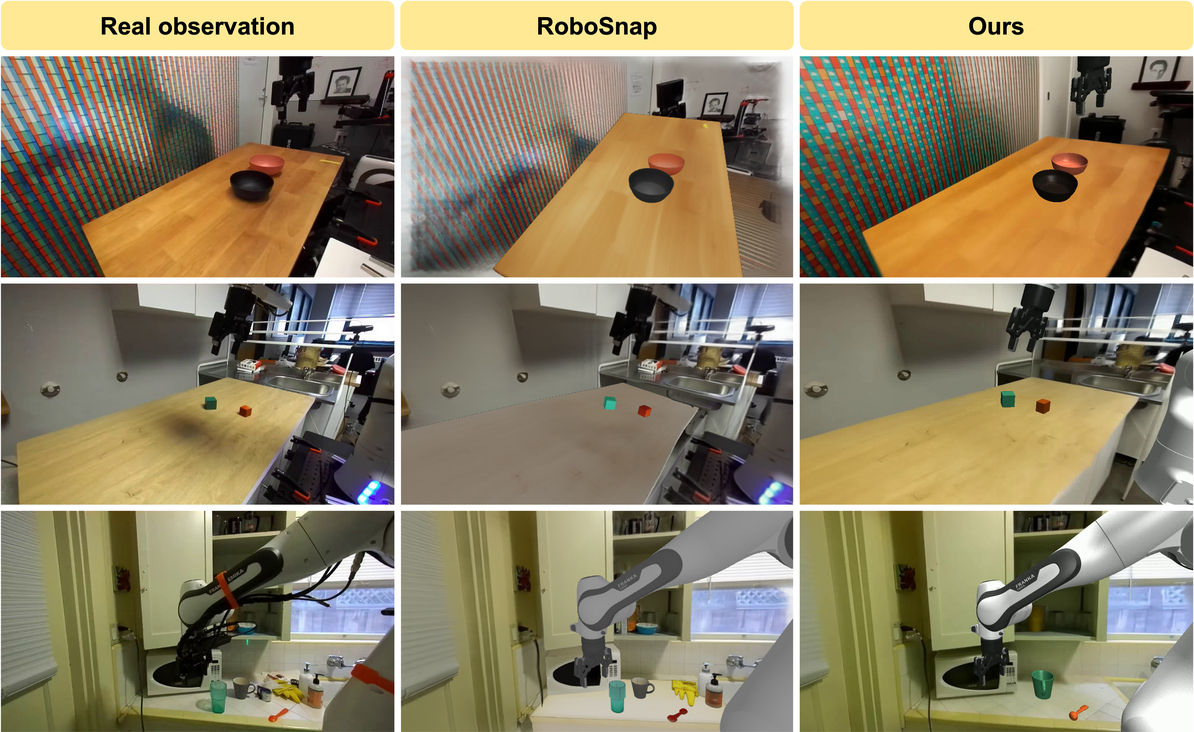}
  \caption{\textbf{Scene reconstruction, ours vs. RoboSnap.}}
  \label{fig:ours-vs-robosnap}
\end{figure}

\paragraph{Qualitative.} Fig.~\ref{fig:droid-qual} pairs real-world observations with their MuJoCo replays for three representative episodes, exposing intermediate pipeline artifacts, e.g., recovered object meshes, FoundationPose-style pose-tracking overlays, and the calibrated robot/ground state. Further, we qualitatively compare our reconstruction of three static scenes from the DROID dataset with the concurrent Robosnap~\cite{zhang2026robosnap}: see Fig.~\ref{fig:ours-vs-robosnap}. Note that our results are comparable to those obtained by Robosnap in terms of visual quality; while for the top and the middle episodes our reconstruction attains slightly better texture and pose estimates, for the bottom scene Robosnap yields closer-to-ground-truth pose and shape estimates of the cup. 

\paragraph{Common Failure Modes.} Reasons for failure includes: objects being nudged, not lifted by the gripper; incomplete segmentation/mesh-recovery (e.g., critical object not segmented, incorrect object chosen for recovery, or a single mesh combining 2+ objects); objects' initial position, scale, or orientation being incorrect, precluding gripping. Factors such as a cluttered scene, occlusion, or low lighting made the conversion less reliable.

\subsection{Multi-VLM Support and Cost Efficiency}

\begin{figure}[t]
  \centering
  \includegraphics[width=\linewidth]{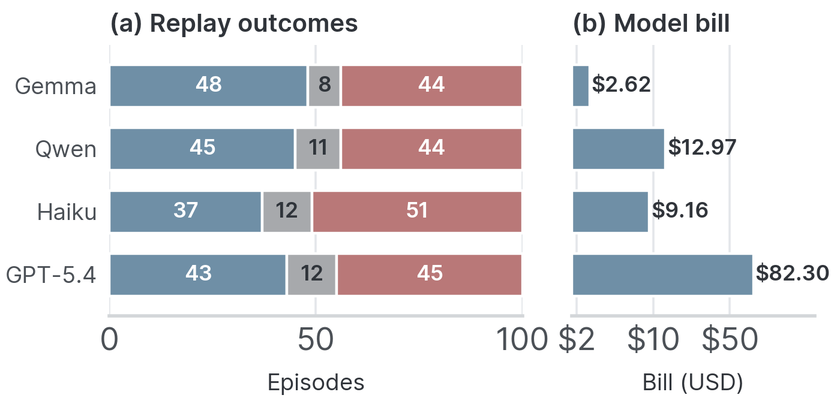}
  \caption{\textbf{Replay outcomes and model bill across VLM backends.} Blue, gray, and red denote replay success ($\geq 8/10$), partial ($7/10$), and failure ($\leq 6/10$ or no valid judge record), respectively. The right panel reports the model bill (USD, log scale) for 100 episodes per backend.}
  \label{fig:droid-vlm-cost}
\end{figure}

Across the four backends assessed, we use the same pipeline configuration, with reasoning enabled for selected agentic steps, and vary only the VLM; all VLM backends are evaluated on the same subset of 100 DROID episodes. The reported model bills include each model's API usage under this configuration. Fig.~\ref{fig:droid-vlm-cost} shows that the four backends have broadly similar observed outcome compositions: Gemma 4 31B, Qwen 3.6 35B, GPT-5.4, and Claude Haiku 4.5 obtain $48/100$, $45/100$, $43/100$, and $37/100$ replay successes, respectively; the numbers are first produced by VLM judges, then further validated by a human assessor. The clearer separation is cost: relative to Gemma's model bill of \$2.62, Claude costs $3.5\times$ as much, Qwen $5.0\times$, and GPT-5.4 $31.4\times$.
 
We attribute the similar outcomes across backends to how the pipeline scopes the VLM's role. A VLM does not generate geometry or simulate physics itself; it simply orchestrates low-level specialized tools for segmentation, stereo depth, pose tracking and deterministic grasp sweep. It is used only for scope-bounded, schema-constrained decisions such as object discovery, keyframe and mask selection under a capped retry budget, and replay-refinement choices. Because these queries are narrowly scoped, they lie well within the capability of modern reasoning VLMs, and the pipeline transfers across backends without requiring per-model tuning: even the 31B open model attains the highest observed success count while using roughly $3\%$ of GPT-5.4's model bill. Backend choice therefore reduces largely to cost, making open-weight models a practical default for large conversion batches. The fact that replay success stays below $50\%$ for every backend further suggests that the remaining headroom lies mostly in the upstream visual and simulation components rather than in the choice of VLM alone.

\paragraph{One-prompt agent baseline.}
These backend comparisons leave open whether Gemma can complete episode conversion without our workflow. We therefore tested Gemma~4 31B in Codex on three DROID episodes previously successfully converted with Gemma and Agentic Real2Sim. Each episode received one independent attempt from an empty Python environment, with its images, robot trajectories, calibration and robot assets, plus a common task prompt and output contract. The agent had up to 60 minutes to install dependencies, call tools and revise its code, without AR2S-specific stages, tools or skills. All three sessions ended before the limit, but none delivered a complete, runnable episode twin. Failures included invalid scene XML, replay errors, stationary task objects and unstable robot states. Together with the backend comparison, these observations support the practical value of a reproducible workflow that works across VLMs, including open-weight models.

\subsection{Deformable and Humanoid Episode Conversion}
\label{sec:exp-deform-humanoid}

\begin{figure}[t]
  \centering
  \includegraphics[width=\linewidth]{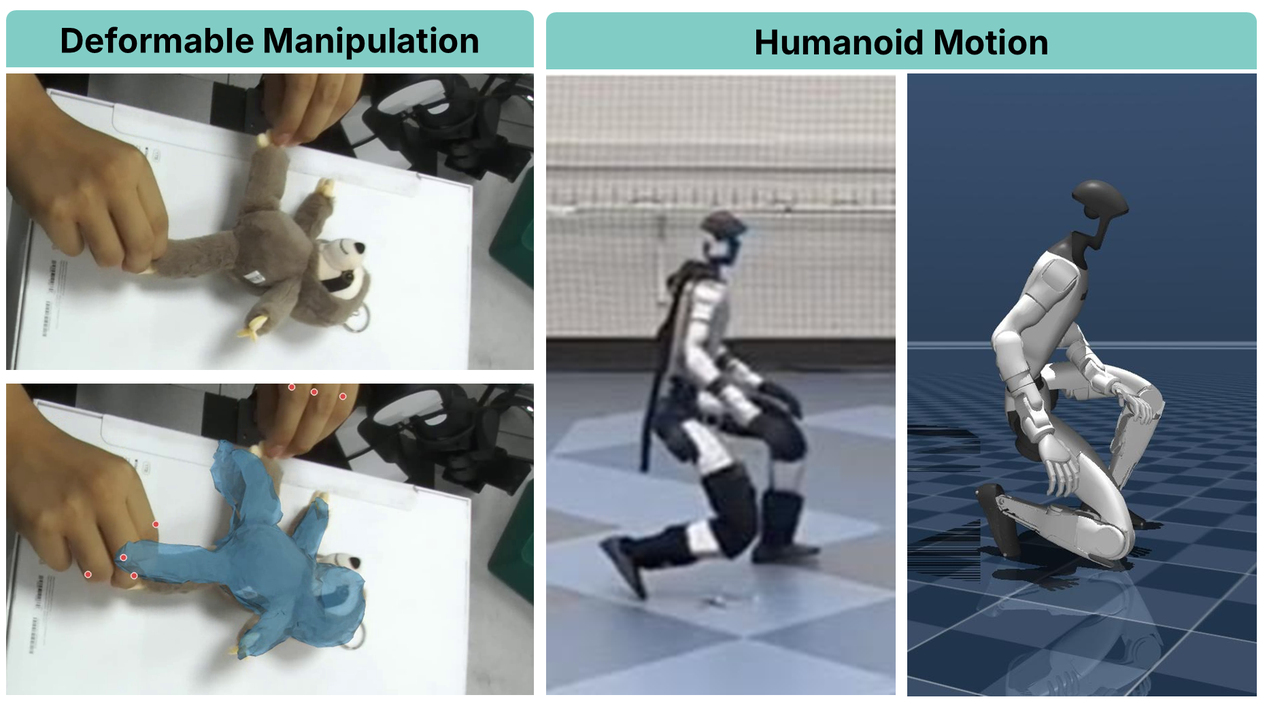}
  \caption{\textbf{Deformable and humanoid episode conversion.} Two non-rigid object manipulation settings that stress the same conversion contract beyond rigid manipulation. \emph{Left:} PhysTwin-style deformable episodes pair real interaction video with the recovered geometry and a deformation overlay. \emph{Right:} Unitree G1 humanoid clips of locomotion and short whole-body motions, retargeted from LAFAN1, where the simulated humanoid is shown against the real frames after the controller's joint-level gains are tuned to match the reference trajectory in closed-loop simulation.}
  \label{fig:deform-humanoid}
\end{figure}

To show that our agentic workflow generalizes to deformables, we convert PhysTwin-style episodes with assets such as rope, cloth, plush objects, soft packages, and elastoplastic materials, and compare the real interaction video with the recovered geometry and simulated deformation. For humanoid motion's twin generation: the input data are Unitree G1 clips of locomotion and short whole-body motions, retargeted from LAFAN1~\cite{harvey2020robust}. We compare the simulated humanoid against the reference motion in closed-loop simulation and report qualitative side-by-side motion comparisons. Fig.~\ref{fig:deform-humanoid} shows both domains. In Fig.~\ref{fig:deform-humanoid} (a), deformable cases pair real interaction video with tracked geometry and a deformation overlay, emphasizing whether the recovered material response follows the visible bending, stretching, and contact sequence. In Fig.~\ref{fig:deform-humanoid} (b), we show a simulated humanoid against its reference motion for standing, kneeling, and short-locomotion clips, emphasizing gross pose, balance, and motion-phase agreement.

\subsection{Robotic Applications}
\label{sec:exp-robotic-applications}

\begin{figure*}[t]
  \centering
  \includegraphics[width=\linewidth]{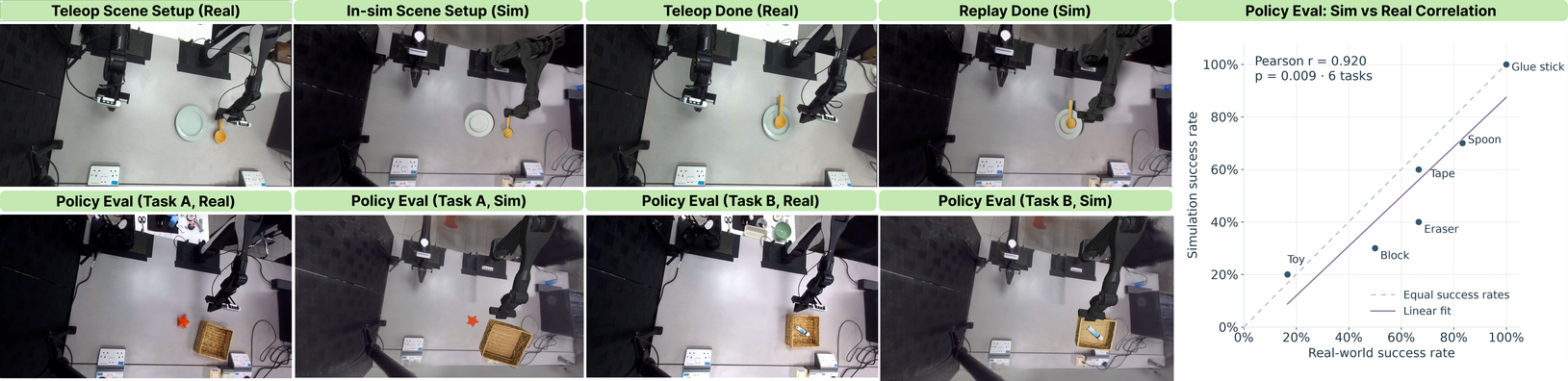}
  \caption{\textbf{Real-world experiments.} \emph{Top:} a teleoperation experiment in which a human operates the AIRBOT Play arm through a VR-headset interface to perform the move-spoon-to-table task, and we reconstruct the episode as a simulated twin. \emph{Bottom:} We evaluate a $\pi_0$ policy \cite{black2024pi0} fine-tuned on real-world data in both real-world tasks and their simulated counterparts. Two tasks are shown here. Success rates in simulation vs. in real world are shown on the right.}
  \label{fig:real-robot}
\end{figure*}

\paragraph{Custom scene conversion.} Beyond converting the DROID recordings, we validate that our framework extends to data collected on a robot platform that does not appear in DROID. Our capture setup consists of an AIRBOT Play arm~\cite{airbot2026}--- a 6-DoF manipulator equipped with the AIRBOT G2 two-finger parallel-jaw gripper, and a single externally mounted ZED~2i stereo camera overlooking the tabletop workspace. Episodes are collected by a human operator who teleoperates the arm through a VR-headset interface (Fig.~\ref{fig:real-robot}, left). The camera records rectified stereo pairs at $2208\times1242$ resolution and $15$\,Hz; arm joint positions and the gripper opening are logged alongside; and the camera-to-robot-base transform is obtained from a one-time station calibration. We package these raw recordings into the DROID episode layout consumed by the pipeline, leaving its ingestion path unchanged; the only embodiment-specific change is a MuJoCo model of the arm and gripper, converted from the vendor-provided URDF and registered with the simulation stage in place of the default DROID Franka model. In the recorded episode, the operator drives the arm to pick a wooden spoon out of a plate resting on the table and set it down on the tabletop beside the plate. Given the packaged episode, the pipeline runs end to end, from object discovery to mesh recovery, followed by scene preparation with simulator-in-the-loop grasp optimization. This yields a MuJoCo twin in which the AIRBOT arm replays the recorded joint trajectory, grasps the spoon, lifts it clear of the plate, and relocates it onto the table, reproducing the outcome of the real episode. Fig.~\ref{fig:real-robot} pairs keyframes of the real recording with the synthetic twin rendered from the calibrated real camera viewpoint.

\begin{figure*}[t]
  \centering
  \includegraphics[width=\linewidth]{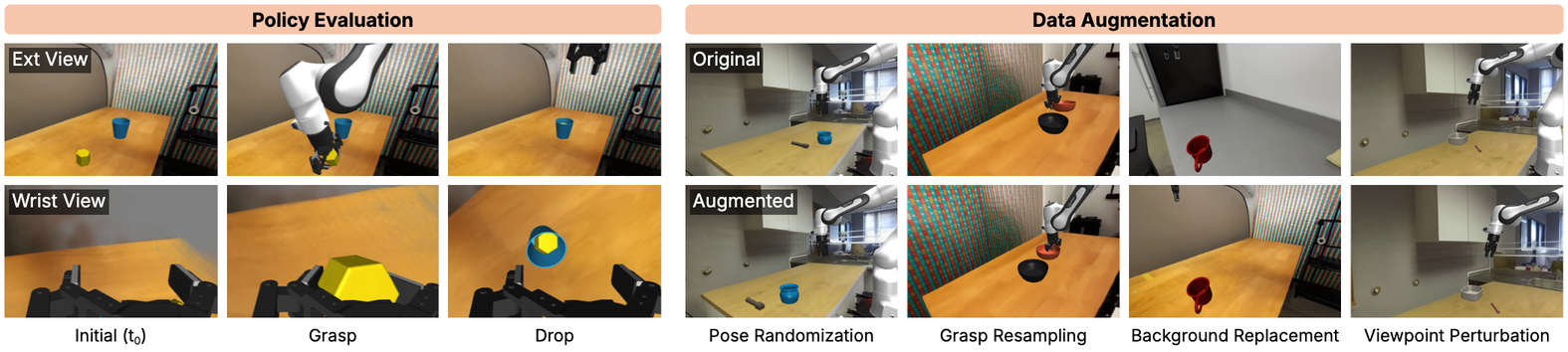}
  \caption{\textbf{Policy evaluation and data augmentation in Agentic Real2Sim episode twins.}
\emph{Left:} Keyframes (initial, grasp, drop) of a closed-loop $\pi_{0.5}$ rollout in a converted
twin, shown from the policy's two composited input views (exterior, top; wrist, bottom). 
\emph{Right:} Augmentations applied to a converted twin (top: original; bottom: augmented), including pose
randomization, grasp resampling, background replacement, and viewpoint perturbation.
Variants are re-simulated and re-rendered inside the twin rather than image-edited, so
every augmented trajectory remains physically and geometrically consistent.}
\vspace{-2mm}
  \label{fig:policy_eval_aug}
\end{figure*}

\paragraph{Data augmentation, in-sim policy training and evaluation.} An episode twin also serves as a basis to generate training data for control policies. Each converted episode includes a grasp demonstration validated through simulator-in-the-loop grasp optimization. Replaying this demonstration in the twin yields synchronized training trajectories comprising composited exterior and wrist observations, proprioceptive states, and actions in the DROID convention, without additional physical data collection. Because the twin exposes the full simulation state and supports rerendering, each demonstration can be expanded through controlled augmentations that are difficult to apply consistently to a fixed real-world recording---see Fig.~\ref{fig:policy_eval_aug} (Right). These augmentations include \emph{object pose randomization}, in which objects are relocated and the end-effector trajectory is recomputed through inverse kinematics; \emph{grasp resampling}, which uses alternative validated grasp points from the grasp sweep to produce distinct demonstrations of the same task; \emph{background replacement}, which substitutes the 3DGS background from another episode; and \emph{camera viewpoint perturbation}, which rerenders both views from perturbed camera poses. Each variant is resimulated and rerendered rather than edited at the image level, preserving physical and geometric consistency. 

The augmented episodes are then used to fine-tune a pretrained $\pi_{0.5}$ policy~\cite{physicalintelligence2025pi05}. In terms of the setup of the evaluation environment, a calibrated MuJoCo arm and gripper are positioned in the recovered scene frame, and reconstructed object meshes are initialized at their grasp-optimized poses. The arm and gripper are then simulated under position control at $15$\,Hz. A wrist-mounted camera, positioned according to the real wrist-camera calibration, completes the DROID camera configuration.  Evaluation-related APIs are exposed through a Gym-style interface, enabling VLA-based manipulation policies to be evaluated with minimal task-specific integration. We evaluate the resulting environment using the DROID-pretrained VLA model $\pi_{0.5}$~\cite{physicalintelligence2025pi05}, configured to predict absolute joint positions. At each control step, the policy receives the composited exterior and wrist RGB views, the robot joint and gripper states, and the episode's natural-language task description. It predicts a chunk of absolute joint-position commands, which the environment executes open-loop before querying the policy again; thus, the rollout is closed-loop at the action-chunk level. Fig.~\ref{fig:policy_eval_aug} (Left) shows a rollout in a converted twin from the policy's two input views: the policy locates and grasps the target object, then drops it into the cup.


\paragraph{Substituting real-world evaluation.}
Using these reconstructed episode twins, we investigate whether simulation can serve as a proxy for real-world policy evaluation. This setting is deliberately separated from the augmentation experiment above: the policy evaluated here does not use synthetic trajectories generated by our framework during training; instead, we fine-tune a $\pi_0$ policy~\cite{black2024pi0} exclusively on real-world demonstrations, and execute the same policy in both a hold-out set of six pick-and-place tasks and their reconstructed simulated counterparts. This allows us to directly assess if policy performance measured in simulation is predictive of its performance in the real robot. Since the policy has never been trained on trajectories generated by the reconstructed twins, agreement between the two domains cannot be attributed to exposure to the simulated evaluation environments during training.

The bottom row of Fig.~\ref{fig:real-robot} shows representative closed-loop policy evaluations for two tasks. Note that the simulated rollout is not a replay of the real policy trajectory: the policy receives observations from the simulated environment and produces actions online, so deviations in visual appearance, object geometry, contact, or robot dynamics can directly affect the resulting task outcome. These deviations may also accumulate over the course of a rollout, making task success a stricter test of the reconstructed environment than matching individual rendered frames or replaying a fixed action sequence.

We evaluate this correspondence across six pick-and-place tasks, and compare task success rates measured in simulation against those measured in the real world. As shown on the right of Fig.~\ref{fig:real-robot}, simulation and real-world performance exhibit a strong Pearson correlation of $r=0.920$ ($p=0.009$). Although the absolute success rates do not exactly match for every task, simulation largely preserves the relative performance across tasks, with easier and harder tasks showing consistent trends in both domains. 

These results indicate that the reconstructed episode twins preserve task-relevant visual and physical factors sufficiently well for closed-loop policy evaluation. In practice, such correspondence enables simulation to act as an intermediate evaluation stage for screening policy variants, identifying promising candidates for deployment, and estimating expected real-world performance before committing to repeated physical trials. 


\section{Conclusions, Limitations and Future Work}
\label{sec:conclusion}

We introduced \textit{Agentic Real2Sim}, a framework for converting real-world physical interaction recordings into simulatable digital twins. The framework preserves the actors, manipulated entities, geometry, physical parameters, camera setup and objects along with end effector trajectories needed for simulation. For the task of converting DROID episodes, Agentic Real2Sim converts robot-object manipulation recordings into MuJoCo episode twins through visual processing, physical-prior inference, scene preparation, and simulator-in-the-loop refinement. The same episode contract also provides a common interface for PhysTwin-style deformable manipulation and BFM-Zero-style humanoid motion, allowing the pipeline to span settings that are usually handled by separate Real2Sim systems. In addition, our framework decouples agentic decisions from any single VLM provider: on DROID-500, an open 31B backend attains comparable observed replay-success outcomes to proprietary alternatives while using only about $3\%$ of GPT-5.4's model cost. Finally, the framework supports custom scene conversion on manipulation data outside the DROID dataset, while the aligned episode twins support downstream policy evaluation and provide generated interaction data for fine-tuning a DROID-pretrained $\pi_{0.5}$ policy~\cite{physicalintelligence2025pi05}.

Our work still has many limitations. First, it focuses on rigid-object DROID episodes, and there is still plenty of room for improvement in terms of successful conversion rate. Second, the framework does not support object-level system identification or directly validate physical parameters, since each converted demonstration provides limited interaction evidence per object. Future work will extend automated conversion to additional deformable episodes and systematically evaluate agentic components and VLM backends for reliability and replay stability.

\section*{Acknowledgment}

This work was supported in part by the NVIDIA Academic Grant Program. We thank NVIDIA for supporting academic research and for providing resources that helped facilitate this work; Wenlong Huang for his help with the DROID dataset; Yiwen Chen for his help with data collection; Beichen Li for his support with computational resources.


\balance
\bibliographystyle{IEEEtran}
\bibliography{main}

\end{document}